\documentclass{article}
\usepackage{spconf,amsmath,graphicx}

\usepackage{multirow}
\usepackage{booktabs}
\usepackage{makecell}
\usepackage{threeparttable}
\usepackage{makecell}
\usepackage{graphicx}

\title{Unimodal Cyclic Regularization for Training Multimodal Image Registration Networks}

\name{Zhe Xu$^{\star \dagger}$ \qquad Jiangpeng Yan$^{\star}$ \qquad Jie Luo$^{\dagger}$  \qquad William Wells $^{\dagger}$ \qquad Xiu Li$^{\star}$ \qquad Jayender Jagadeesan$^{\dagger}$}
\address{$^{\star}$ Shenzhen International Graduate School, Tsinghua University, Shenzhen, China \\
    $^{\dagger}$ Brigham and Women’s Hospital, Harvard Medical School, Boston, USA}
\begin{document}
%
\maketitle
\begin{abstract}
The loss function of an unsupervised multimodal image registration framework has two terms, i.e., a metric for similarity measure and regularization. In the deep learning era, researchers proposed many approaches to automatically learn the similarity metric, which has been shown effective in improving registration performance. However, for the regularization term, most existing multimodal registration approaches still use a hand-crafted formula to impose artificial properties on the estimated deformation field. In this work, we propose a unimodal cyclic regularization training pipeline, which learns task-specific prior knowledge from simpler unimodal registration, to constrain the deformation field of multimodal registration. In the experiment of abdominal CT-MR registration, the proposed method yields better results over conventional regularization methods, especially for severely deformed local regions.

\end{abstract}
\section{Introduction}
\label{sec:intro}
Medical image registration is an essential procedure in many image-guided therapies. With the advent of deep learning (DL) techniques, the development of image registration algorithms have moved towards a learning-based framework  \cite{sedghi2019semi, hu2018weakly, cao2018deep}. One promising framework is the unsupervised image registration \cite{VM2018, xu2020f3rnet, Xu2020unsupervised}, where a neural network is trained to estimate a deformation field that minimizes a loss function using unlabeled image pairs. 

In learning-based unsupervised image registration, choosing the appropriate loss function is the key to achieving accurate results. A loss function contains two terms, i.e., a metric for similarity measure, and regularization. Conventionally, approaches for unimodal registration use intensity-based similarity metrics such as mean squared error (MSE), whereas approaches for multimodal registration use more complex metrics, such as mutual information \cite{wells1996multi} and MIND \cite{mind}. For the regularization term, most existing approaches use hand-crafted formula, i.e., L1 or L2-norm smoothness, to impose artificial properties to the estimated deformation field (DF).

Distinct from using aforementioned hand-crafted metrics, researchers have proposed approaches to automatically learn the similarity metric, which have been shown effective in improving the registration performance. However, less attention has been paid to automatically learning the regularization term. 

\begin{figure}[htb]
  \centering
  \vspace{-0.2cm}
  \centerline{\includegraphics[width=8.5cm]{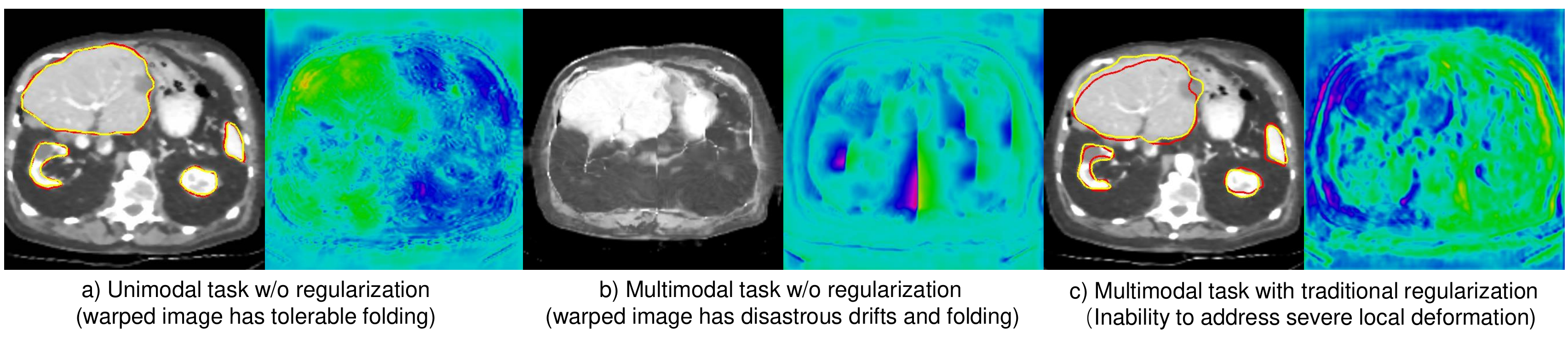}}
  \vspace{-0.2cm}
  \caption{ Illustration for the influence of regularization in medical image registration.}
  \vspace{-0.2cm}
  \label{fig_int}
\end{figure}

Aligning abdominal CT to MR is a challenging multimodal registration problem. Due to its complexity, multimodal registration may rely on the regularization term to constrain the DF more than unimodal registration does. In an exploratory study, we have observed that even without a regularization term, a unimodal registration network is still able to achieve a satisfactory result using MIND \cite{mind} similarity metric, which can be seen in Fig.\ref{fig_int}(a). Yet for multimodal registration, as shown in Fig.\ref{fig_int}(b), the network tends to overfit and produce chaotic deformation fields without regularization. 

One disadvantage of conventional hand-crafted regularization formulas is that they apply generic constrains to all regions across the image. As shown in Fig.\ref{fig_int}(c), the universal smoothness constraint is not ideal for abdominal CT to MR registration because some organs, i.e., lobes of the liver with cancer, experience severe deformation due to progressed disease or insufflation during surgery than organs in other regions of the image. These organs can not be properly registered if the same universal regularization term is applied. Therefore, developing a task-specific regularization term that accounts for the strength of deformation in different image regions can considerably improve the performance of CT to MR registration. 

Recently, adversarial deformation regularization \cite{hu2018adversarial} was proposed. Within their semi-supervised MR-TRUS registration pipeline, an additive discriminator network was trained as a regularization term to distinguish the predicted DFs from pre-generated finite element analysis motion simulated DFs. However, the finite element analysis based biomechanical simulation is computationally expensive and difficult to acquire. 

In this work, we propose a unimodal  cyclic  regularization training pipeline, which learns task-specific prior knowledge from simpler unimodal registration, to constrain the deformation field of multimodal registration. Take CT-to-MR registration image for example, by first utilizing traditional registration algorithms to form a pre-registered dataset, a unimodal registration network can be pre-trained to build the backward mapping between the warped CT images and the original moving CT images for multimodal registration. As a result, the inverse multimodal DFs can be estimated by integrating the simpler unimodal registration model, where it bridges the gap between uni- and multi-modal registration. In other words, the model-based backward mapping has learned task-specific biologically-plausible prior knowledge during the pre-training process, which can better regularize multimodal registration for better alignment with respect to local regions while guaranteeing global fidelity. The method is evaluated on a clinically acquired abdominal CT-MR dataset. We show that the proposed pipeline can achieve better performance compared to other competing conventional approaches.

\section{Methods}
\label{sec:methods}
\subsection{Cyclic Regularization Training Pipeline}
\subsubsection{Forward-Multimodal Registration Network}

\begin{figure}[htb]
  \centering
  \centerline{\includegraphics[width=7.5cm]{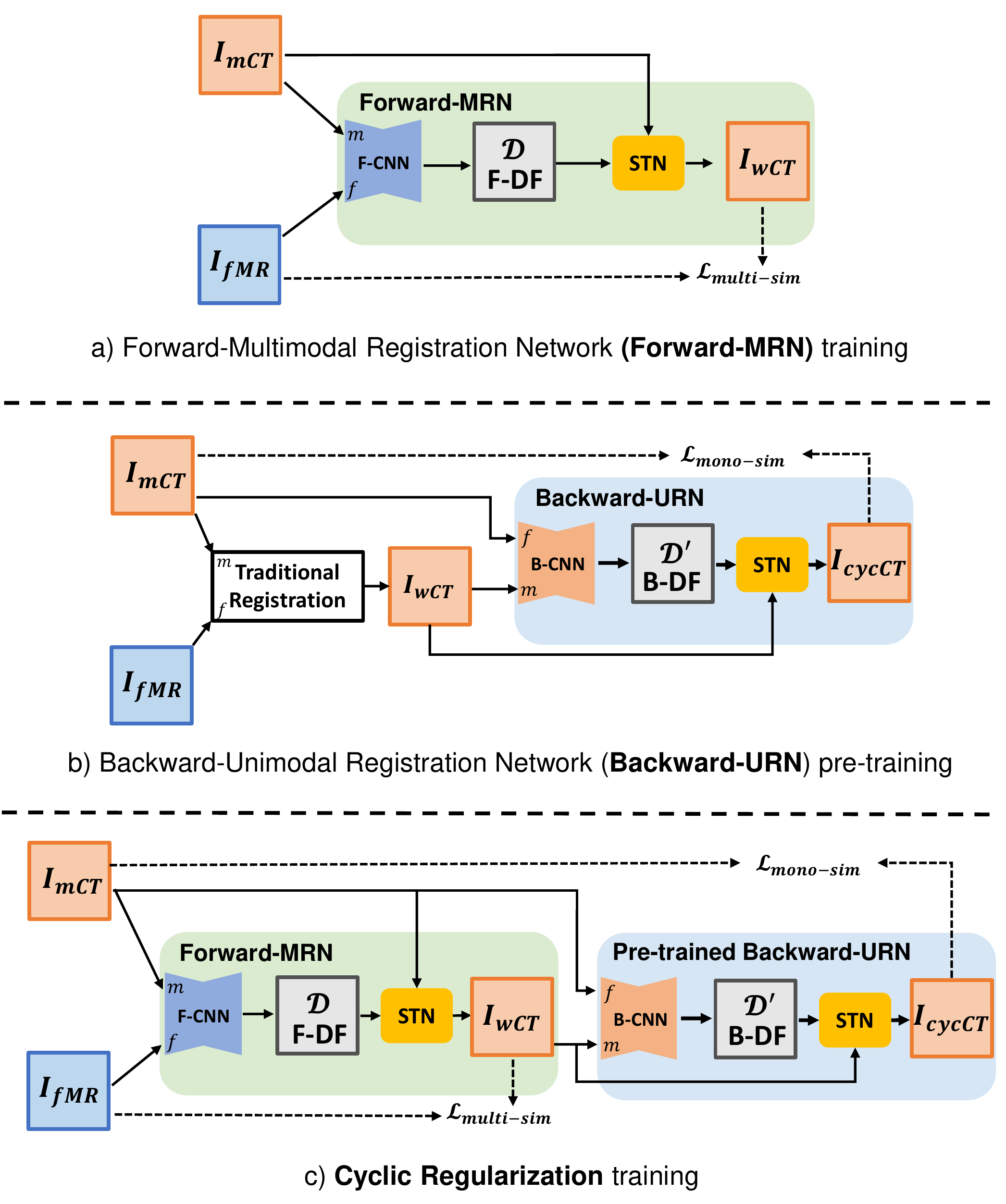}}
  \caption{Illustration of our proposed cyclic regularization training pipeline.}
  \vspace{-0.4cm}
  \label{fig2}
\end{figure}

Fig.~\ref{fig2} (a) illustrates the training pipeline for a standard unsupervised multimodal image registration network, which we name it as Forward-Multimodal Registration Network (Forward-MRN). Given a moving CT image $I_{mCT}$ and a fixed MR image $I_{fMR}$, in most unsupervised multimodal image registration approaches, the estimated deformation field $\mathcal{D}$ is obtained by optimizing the following loss function: 
\begin{small}
\begin{equation}
\mathcal{L}\left(I_{f M R}, I_{m C T}, \mathcal{D}\right)=\mathcal{L}_{m u l t i-s i m}\left(I_{f M R}, I_{m C T} \circ \mathcal{D}\right)+\alpha \mathcal{L}_{r e g},
\end{equation}
\end{small}
where $\mathcal{L}_{multi-sim}$ measures the image (dis)similarity between the fixed image $I_{fMR}$ and the warped image $I_{mCT} \circ \mathcal{D}$. Here $\mathcal{L}_{r e g}$ represents the regularization term, and $\alpha$ is a weight parameter. In most existing approaches, $\mathcal{L}_{r e g}$ adopts hand-crafted L1/L2-norm smoothness, bending energy, etc. 

\subsubsection{Backward-Unimodal Registration Network} The proposed task-spesific regularization term is based on a cyclic transformation of $I_{mCT}$ with the Backward-Unimodal Registration Network (Backward-URN), whose pipeline is shown in Fig.1 (b). Backward-URN is designed to obtain the deformation field $\mathcal{D}^{\prime}$ for recovering moving CT images ($I_{mCT}$) from warped CT images ($I_{wCT}$) under the supervision of a unimodal image (dis)similarity metric. We pre-train the Backward-URN using a dataset of paired $I_{mCT}$ and $I_{wCT}$, which are pre-registered by traditional multimodal registration algorithms. It is noteworthy that some unimodal registration researches provide off-the-shelf trained models, e.g., \cite{VM2018} for unimodal brain registration. It is possible to fine-tune their models if the registration is for brain scans. This work provides a general pipeline for various tasks since many traditional methods are easily accessible. The remaining procedure follows the standard unsupervised unimodal registration training by optimizing:
\begin{small}
\begin{equation}
\begin{array}{l}
\mathcal{L}\left(I_{m C T}, I_{w C T}, \mathcal{D}^{\prime}\right)=\mathcal{L}_{m o n o-s i m}\left(I_{m C T}, I_{w C T} \circ \mathcal{D}^{\prime}\right) \\
{\kern 88.3pt} +\beta \mathcal{L}_{s m o o t h}.
\end{array}
\vspace{-0.3cm}
\end{equation}
\end{small}

Here $\mathcal{L}_{mono-sim}$ is the unimodal image (dis)similarity between $I_{cycCT}$ and $I_{mCT}$, $\mathcal{L}_{smooth}$ represents smoothness penalty for $\mathcal{D}^{\prime}$, $\beta$ is a regularization parameter. It is noteworthy that the traditional smoothness regularization is only used for pre-training the backward-URN as it is sufficient for regularizing the simpler unimodal registration task.

\vspace{-0.1cm}
\subsubsection{Cyclic Regularization Training} 
As shown in Fig.~\ref{fig2}(c), in order to take advantage of the learned prior knowledge, we cascade the fixed-weight pre-trained Backward-URN with Forward-MRN, and use unimodal image similarity $\mathcal{L}_{m o n o-s i m}$ between $I_{mCT}$ and $I_{cycCT}$($I_{w C T} \circ \mathcal{D}^{\prime}$) as the regularization term $\mathcal{L}_{r e g}$ to optimize Forward-MRN. Therefore, the loss function of our cyclic regularization training can be defined as:
\begin{small}
\begin{equation}
\label{equ_final}
\begin{array}{l}
\mathcal{L}\left(I_{f M R}, I_{m C T},I_{w C T}, \mathcal{D},\mathcal{D}^{\prime}\right)=\mathcal{L}_{m u l t i-s i m}\left(I_{f M R}, I_{m C T} \circ \mathcal{D}\right)\\
{\kern 106.3pt}+\alpha \mathcal{L}_{m o n o-s i m}\left(I_{m CT}, I_{w C T} \circ \mathcal{D}^{\prime}\right).
\end{array}
\end{equation}
\end{small}


As mentioned above, the deformation field $\mathcal{D}$ is used to register $I_{mCT}$  to $I_{fMR}$ under the supervision of multimodal image similarity, while the deformation field $\mathcal{D}^{\prime}$ is used to deform $I_{wCT}$ back to $I_{mCT}$ using unimodal image similarity. Since the fixed-weight Backward-URN and the Spatial Transformation Networks (STN) \cite{STN} do not have trainable parameters during the cyclic regularization stage, the gradients can be directly back-propagated to optimize the Forward-MRN, which can implicitly regularize $\mathcal{D}$ to be plausible and smooth. As such, the model-based regularization with biologically-plausible prior knowledge can be more capable of handling large local deformations. For example, in a CT-MR abdominal registration where livers are severely deformed, conventional hand-crafted regularization formulas tend to sacrifice the local alignment to maintain satisfactory overall registration accuracy, while the proposed regularization can be better at aligning severely deformed local regions while ensuring the fidelity of other surrounding organs.

The detailed training strategies are shown in section \ref{trainingStragegy}. After cyclic regularization training, we can efficiently use Forward-MRN for CT-to-MR image registration.


\subsection{Network Architectures}
\begin{figure}[htb]
  \centering
  \centerline{\includegraphics[width=8.5cm]{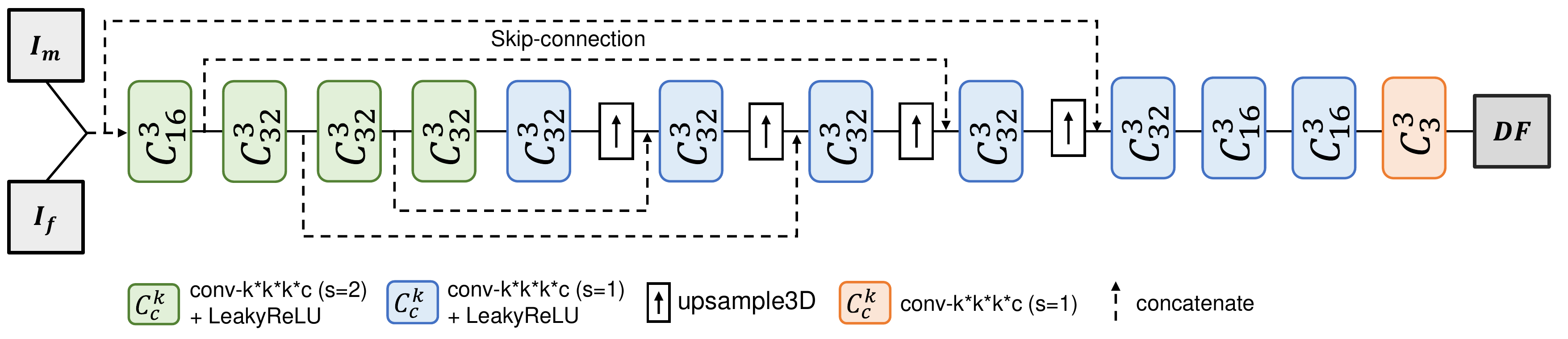}}
  \caption{Detailed architecture of F-CNN and B-CNN.}
  \vspace{-0.6cm}
  \label{fig3}
\end{figure}

The F-CNN and B-CNN within Forward-MRN and Backward-URN both adopt the same CNN architecture introduced in VoxelMorph \cite{VM2018}. As shown in Fig.\ref{fig3}, the moving image $I_{m}$ and the fixed image $I_{f}$ are concatenated as a 2-channel input and downsampled by four $3 \times 3 \times 3 $ convolutional layers with stride of 2 as the encoder module. The network is then upsampled four times accordingly to recover the size of feature maps as the decoder module. Skip connections between the encoder and the decoder are also applied. Another four convolutional layers are adopted to refine the number of feature maps and produce the final DF with 3 channels. After estimating the DF, the spatial transformer is applied to warp $I_{m}$.

\subsection{Loss Function}
Our cyclic regularization pipeline consists of two unsupervised registration networks: 1) pre-trained Backward-URN and 2) Forward-MRN. 

Particularly, the similarity metrics for both networks use the same Modality Independent Neighborhood Descriptor (MIND) \cite{mind} since it is a structural representation that is invariant to different modalities. Using MIND-based similarity in both networks makes it easier to determine the weight of $\mathcal{L}_{mono-sim}$ as the value ranges between $\mathcal{L}_{mono-sim}$ and $\mathcal{L}_{multi-sim}$ are similar. MIND-based metric can be defined as:
\begin{small}
\begin{equation}
\mathcal{L}_{M I N D}\left(I_{w}, I_{f}\right)=\frac{1}{N|R|} \sum_{x}\left\|M I N D\left(I_{w}\right)-M I N D\left(I_{f}\right)\right\|_{1},
\end{equation}
\end{small}
where $N$ denotes the number of voxels in $I_{w}$ and $I_{f}$, $R$ is a non-local region around voxel $x$. 

For the regularization term, Backward-URN uses the traditional L2-norm of DF gradients, while Forward-MRN uses the proposed unimodal cyclic regularization.


\section{Experiments and Results}
\subsection{Dataset and Training Strategies}
\label{trainingStragegy}
We focus on the application of abdominal CT-MR registration. Under the IRB approved study, we obtained an intra-patient CT-MR dataset that was collected from 50 patients. The liver, kidney and spleen are manually segmented for quantitative evaluation. All the images were resampled to the same resolution (${1mm}^{3}$), and affine spatial normalization of CT and MR images was performed using ANTs Toolkit \cite{Avants2011ARE}. The images are normalized and cropped to $176\times176\times128$. 

In the pre-training stage, we randomly chose 25 pairs of CT and MR images and used two conventional multimodal registration approaches, ElasticSyN and SyN \cite{Avants2008SymmetricDI}, to align CT images onto MR images. Consequently, a new 50-pair pre-training set with traditional warped CT images ($I_{wCT}$) and original moving CT images ($I_{mCT}$) was formed. To enhance the generalizability of Backward-URN, $I_{wCT}$ and $I_{mCT}$ were regarded as moving images in turn. The weight of the L2-norm smoothness term was empirically set to 0.5. 

After pre-training the Backward-URN, we further selected 15 more pairs randomly to form a 40-pair training set for cyclic regularization training, and the remaining 10 pairs were used for testing. After grid search, the weight of the backward regularization was set to 0.5. 

The method was implemented on Keras with the Tensorflow backend and trained on an NVIDIA Titan X (Pascal) GPU. We adopted Adam as optimizer with a learning rate of 1e-5 and the batch size of 1. 

\begin{table*}[]
\caption{Comparison of Dice Score and ASD on different regularization methods.}\label{tab1}
\centering
\scalebox{0.75}{
\begin{tabular}{p{1.0cm}<{\centering}|p{1.0cm}<{\centering}|p{2.0cm}<{\centering}|p{2.0cm}<{\centering}|p{2.1cm}<{\centering}|p{2.0cm}<{\centering}|p{2.0cm}<{\centering}|p{2.0cm}<{\centering}|p{2.0cm}<{\centering}}
\Xhline{1pt}
\textbf{Metric}                   & \textbf{Organ} & \textbf{Moving} & \textbf{SyN} & \textbf{\begin{tabular}[c]{@{}l@{}}w/o Reg\end{tabular}} & \textbf{\begin{tabular}[c]{@{}l@{}}L2\end{tabular}} & \textbf{\begin{tabular}[c]{@{}l@{}}L1\end{tabular}} & \textbf{\begin{tabular}[c]{@{}l@{}}BE\end{tabular}} & \textbf{Ours} \\ \hline
\multirow{3}{*}{\textbf{\begin{tabular}[c]{@{}c@{}}Dice\\ (\%)\end{tabular}}} & Liver  &       76.23 $\pm$4.48          &       79.42$\pm$4.25       &      45.37$\pm$7.69                                                                 &          82.97$\pm$3.73                                                   &             81.03$\pm$3.55                                                &           83.05$\pm$3.28                                                      &      \textbf{86.67$\pm$3.07}         \\ \cline{2-9} 
                                   & Spleen &        77.94$\pm$3.49         &    80.33$\pm$3.37          &       36.24$\pm$6.23                                                                &   82.28$\pm$3.02                                                          &       81.84$\pm$3.14                                                      &      83.46$\pm$2.95                                                           &       \textbf{85.13$\pm$2.99}        \\ \cline{2-9} 
                                   &Kidney &       80.18$\pm$3.24          &      82.68$\pm$2.96        &            42.86$\pm$7.42                                                           &     85.01$\pm$3.18                                                        &          84.59$\pm$3.07                                                   &   84.67$\pm$3.52                                                              &     \textbf{85.38$\pm$3.11}          \\ \hline
\multirow{3}{*}{\textbf{\begin{tabular}[c]{@{}c@{}}ASD\\ (mm)\end{tabular}}}  & Liver  &       4.98$\pm$0.85          &       4.83$\pm$0.73       &    8.42$\pm$2.18                                                                   &        3.94$\pm$0.64                                                     &          3.82$\pm$0.69                                                   &              3.52$\pm$0.82                                                   &   \textbf{2.69$\pm$0.56}            \\ \cline{2-9} 
                                   & Spleen &        2.02$\pm$0.68         &        1.62$\pm$0.62      &     7.04$\pm$1.97                                                                  &    1.45$\pm$0.59                                                         &                                                   1.51$\pm$0.54         &              1.48$\pm$0.61                                                   &  \textbf{1.31$\pm$0.57}             \\ \cline{2-9}
                                   & Kidney &        1.95$\pm$0.43         &        1.91$\pm$0.47      &     6.83$\pm$2.09                                                                  &    1.73$\pm$0.38                                                         &                                                    1.78$\pm$0.33         &              1.69$\pm$0.37                                                   &  \textbf{1.45$\pm$0.39}
                                   \\ \Xhline{1pt}
\end{tabular}}
\vspace{-0.3cm}
\end{table*}

\subsection{Experimental Results}
\subsubsection{Cyclic registration within our pipeline}
\begin{figure}[htb]
  \centering
  \centerline{\includegraphics[width=8.2cm]{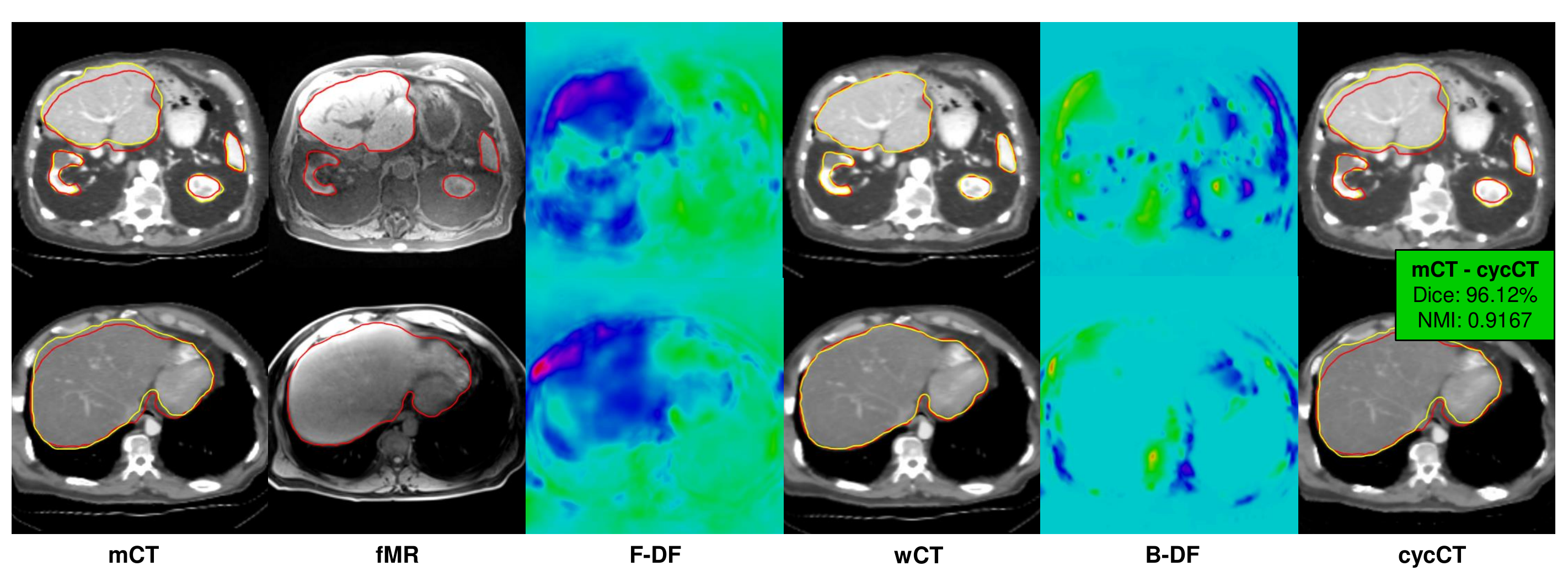}}
  \caption{Example results within our pipeline.}
  \label{fig4}
\end{figure}

We visualize two registered image examples and corresponding deformation fields (F-DF and B-DF) using Forward-MRN and Backward-URN in Fig.\ref{fig4}. We can observe that our Forward-MRN can effectively align the moving CT (mCT) images to the fixed MR (fMR) images, while the Backward-URN is also capable of estimating an approximate inverse deformation to align the warped CT (wCT) images back to the original mCT with 96.12\% of average Dice scores and 0.9167 of normalized mutual information (NMI) between cycCT and mCT.

\subsubsection{Comparisons with other state-of-the-art approaches}
The traditional method \textbf{SyN (MI)} \cite{Avants2008SymmetricDI} is used as the baseline. Besides, our proposed method is compared with three classical regularization methods, including \textbf{bending energy (BE)}, \textbf{L1-norm} and \textbf{L2-norm} of deformation field gradients, with the same backbone network and MIND-based similarity metric as in Forward-MRN. We trained these competing methods with six different weights \{0.1, 0.5, 1, 1.5, 2, 2.5\}, and adopted the optimal weighting value that produces the highest average Dice score over ROIs for each method. Apart from the regularization term, all the networks were subjected to the same training procedure. From the reported results, the weights of L1-norm, L2-norm, and bending energy are set to 1, 1.5 and 2, respectively. When testing on an image pair, SyN costs more than 5 minutes, while learning-based methods can complete registration in less than a second with a GPU. 
\begin{figure}[htb]
  \centering
  \centerline{\includegraphics[width=8.5cm]{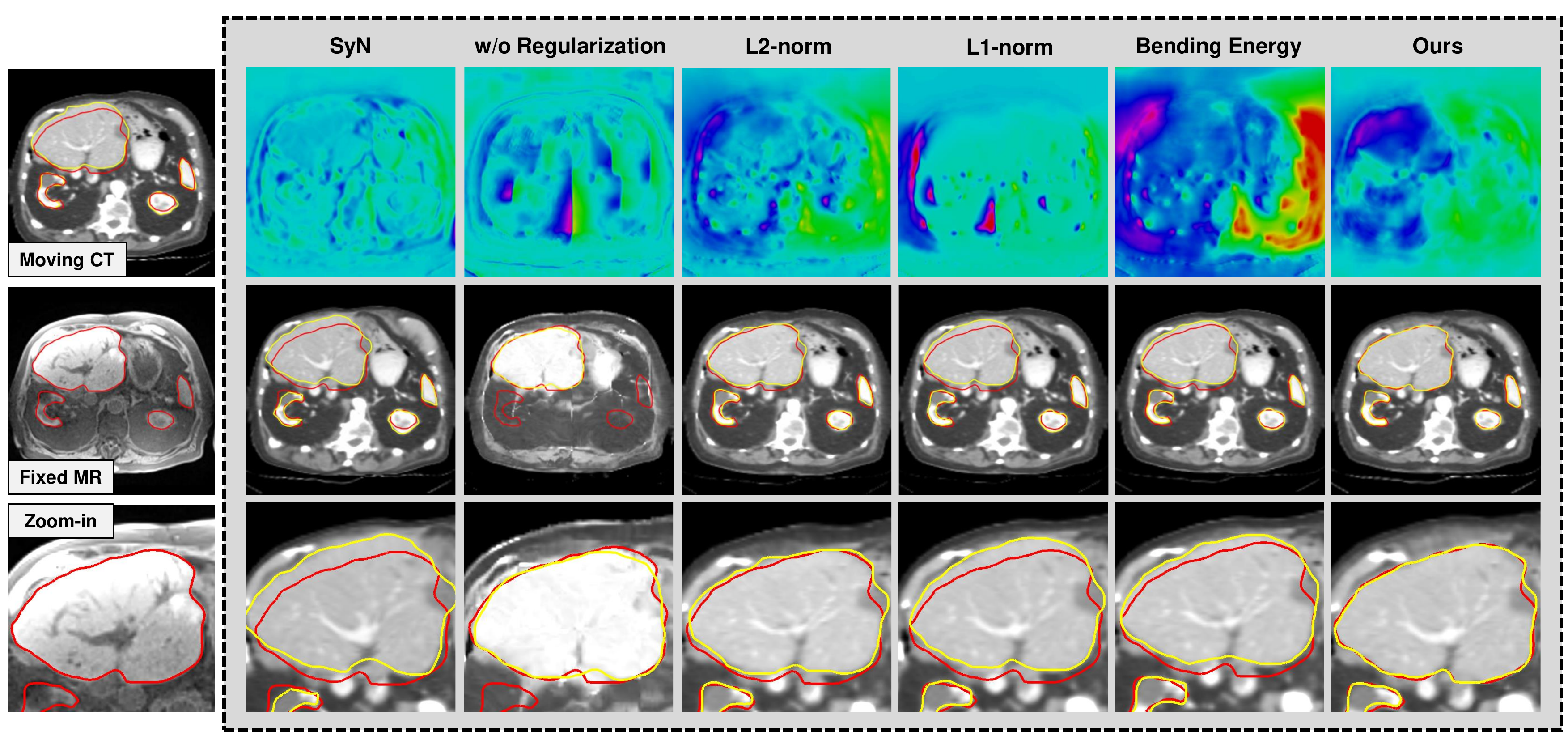}}
  \caption{The left column shows an example moving CT image and fixed MR image. The corresponding warped CT images and deformation fields with different regularizations are presented in the right box. The red contours represent ground-truth segmentation in the fixed MR image while yellow shows the warped segmentation in the corresponding CT image.}
  \label{fig5}
\end{figure}

Fig.5 illustrates an example of registration results produced by our method and other methods. As we have mentioned above, liver registration is much more challenging in abdominal registration task. From the results, we can see that our method can better align the liver with large local deformation, which shows that the forward multimodal registration exactly benefits from the task-specific backward regularization employing simpler unimodal registration.

Table~\ref{tab1} provides the Dice score and Average Surface Distance (ASD) for all competing approaches. Consistent with visual results, the proposed method achieves a significantly higher average Dice score and lower ASD than the same networks trained with L1-norm, L2-norm and bending energy, especially in liver registration. 

\section{Conclusion}
In this work, we propose a novel training pipeline for unsupervised multimodal deformable registration, which incorporates a task-specific unimodal cyclic regularization instead of traditional deformation smoothness. Experimental results indicate that our method can more accurately deal with large local deformations, and it outperforms multiple baselines with other traditional regularizations. Incorporating regularization with more prior knowledge may become a future direction as it is essential for many disease-specific applications. 

\section{Compliance with Ethical Standards}
All procedures performed in studies involving human participants were in accordance with the ethical standards of the institutional and/or national research committee and with the 1964 Helsinki Declaration and its later amendments or comparable ethical standards.

\section{Acknowledgements}
This project was supported by the National Institutes of Health (Grant No. R01EB025964, R01DK119269, and P41EB015898), the National Key R\&D Program of China (No. 2020AAA0108303), NSFC 41876098 and the Overseas Cooperation Research Fund of Tsinghua Shenzhen International Graduate School (Grant No. HW2018008).

\bibliographystyle{IEEEbib}
\bibliography{strings,refs}

\end{document}